\documentclass[conference,a4paper]{IEEEtran}
\IEEEoverridecommandlockouts

\usepackage[hidelinks]{hyperref}
\usepackage[cmex10]{amsmath}%American Math Society(AMS) math formatting
\usepackage{amssymb,amsfonts}%AMS extra symbols and fonts
\usepackage{dblfloatfix}%fix double column figure ordering and placement

\usepackage[ruled,vlined]{algorithm2e}
\usepackage{graphicx}
\graphicspath{{Figures/PDF/}{Figures/PNG/}}

\usepackage{booktabs}
\usepackage{siunitx}
\usepackage[numbers,compress]{natbib}
\usepackage{texnames}
\usepackage{bm,bbm}
\usepackage{orcidlink}
\usepackage{amsmath,amsfonts,amssymb}
\usepackage{multirow, makecell}
\usepackage{subfig}

\begin{document}

\title{ELMZIP: ONBOARD SATELLITE IMAGE COMPRESSION VIA EXTREME LEARNING MACHINES FOR EFFICIENT DOWNLINK}

\author{\IEEEauthorblockN{Woojin Cho \;\; Junghwan Park \;\; Sangcheol Sim \;\; Steve Andreas Immanuel \;\; Junhyuk Heo \;\; Darongsae Kwon}
\IEEEauthorblockA{
TelePIX \\
07330, Seoul, South Korea
\\
\{woojin, junghwan, sim2real, steve, hjh1037, darong.kwon\}@telepix.net \\
}
}

\maketitle
\begin{abstract}
The acquisition of multispectral imagery via small satellites (e.g., CubeSats) presents significant data downlink challenges due to high data volumes and restricted communication windows. While onboard image compression is critical to address this bottleneck, traditional methods often struggle to adapt to the nonlinear statistics of multi-band, multi-resolution data. To overcome these limitations, we propose ELMZip, a novel framework based on Extreme Learning Machines (ELM) and domain decomposition strategies for efficient, resolution-free onboard neural representation.
ELMZip formulates the fitting process as a convex least-squares problem using random-feature single-layer networks, thereby eliminating the need for computationally expensive backpropagation. By adopting an asymmetric transmission protocol that sends only the compact output weights, the proposed method significantly reduces the downlink payload. Unlike previous neural representation approaches that rely on iterative optimization and require transmitting full network parameters, ELMZip achieves significant compression efficiency while maintaining high reconstruction fidelity. This capability enables immediate image reconstruction for analysis, allowing resource-constrained platforms to maximize data return and advancing real-time AI-powered Earth observation.
\end{abstract}


\section{Introduction}

Small satellites such as CubeSats are transforming Earth observation, yet their onboard computing and communication capabilities remain tightly constrained by limited power budgets, passive thermal management, radiation effects, and short downlink windows. As sensor performance improves, the resulting growth in high-resolution multispectral data increasingly exceeds available satellite-to-ground bandwidth, forcing operators to discard data, delay transmission, or rely on conventional compression standards such as CCSDS 123.0~\cite{hernandez2021ccsds} and JPEG~\cite{wallace1992jpeg}. While these codecs are effective, their fixed transforms may not fully adapt to the highly nonlinear, scene-dependent statistics of diverse land covers.

Motivated by these operational constraints, onboard edge computing has gained momentum with the availability of low-power AI hardware (e.g., the NVIDIA Jetson series)~\cite{mittal2019survey, karumbunathan2022nvidia, buonaiuto2017satellite, giuffrida2021varphi}, enabling more flexible and task-aware data reduction directly on spacecraft. In parallel, implicit neural representations (INRs) have been explored as an alternative data representation for compression by modeling imagery as a continuous function over spatial coordinates rather than a discrete pixel grid~\cite{cho2025neural, chen2022videoinr}. Despite strong reconstruction quality, widely used INR architectures such as SIREN~\cite{sitzmann2020implicit} or WIRE~\cite{saragadam2023wire} typically require iterative backpropagation, leading to substantial latency and energy consumption that are difficult to accommodate during time-limited orbital passes. Moreover, transmitting full network parameters can diminish the practical compression gains, limiting their utility for bandwidth-constrained missions.

% In this work, we propose ELMZip, an Extreme Learning Machine–based neural compression framework for rapid, image regression and efficient downlink of multispectral imagery. By leveraging fixed random features and solving only the output weights via a convex least-squares formulation, ELMZip enables fast onboard fitting and transmits a compact set of output parameters for reconstruction at the ground station.

\begin{figure}[t]
\centering
\includegraphics[width=1.0\columnwidth]{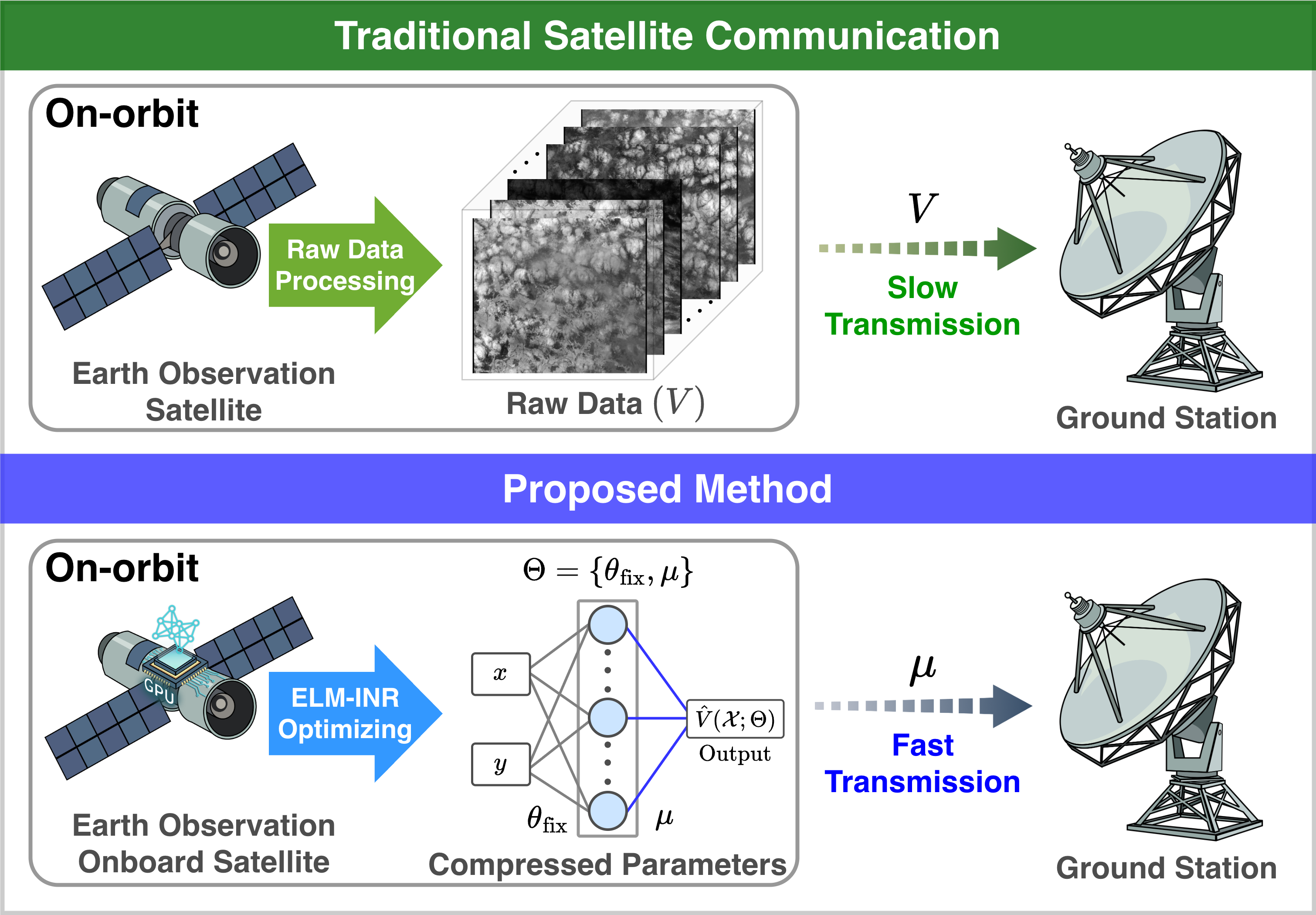}
\caption{Traditional satellite communication transmits raw image data, whereas the proposed method transmits compact neural representation parameters for faster ground reconstruction.}
\label{fig:concept_vis}
\end{figure}

In this work, we propose Extreme Learning Machine–based neural compression (ELMZip), a novel framework designed specifically for onboard satellite image compression. Drawing inspiration from domain decomposition-based ELM architectures~\cite{anderson2024elm, van2026local} used for solving partial differential equations, we adapt the principles of domain decomposition and Extreme Learning Machines (ELM)~\cite{huang2006extreme} to the problem of multispectral image regression. Unlike conventional deep learning methods, ELMZip reformulates per-image training as a convex linear least-squares problem, enabling an analytic solution in seconds without iterative backpropagation.

\begin{figure*}[ht!]
\centering
\includegraphics[width=1.9\columnwidth]{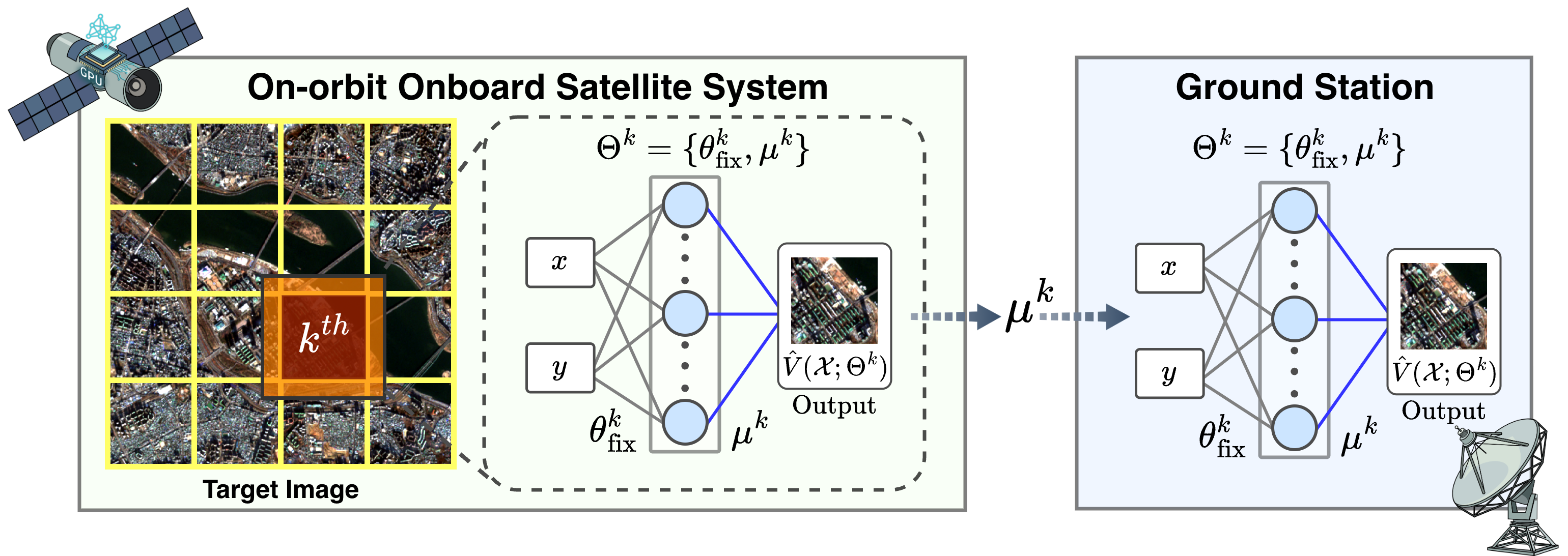}
\caption{\textbf{Overview of ELMZip}: the satellite fits only the output weights $\mu^k$ using fixed random features $\theta_{\text{fix}}$ pre-shared between the satellite and the ground station, and transmits only the $\mu^k$, enabling the ground station to reconstruct $\hat{V}(\mathcal{X};\Theta^k)$.}
\label{fig:model_archi}
\end{figure*}

To maximize downlink efficiency, ELMZip adopts an asymmetric transmission protocol. By synchronizing the random feature initialization between the satellite and the ground segment (via a shared deterministic initialization), the satellite needs to transmit only a compact set of output parameters (i.e., the final-layer weights), rather than the full network parameters as in standard INR pipelines~\cite{cho2025fourier}. As a result, the ground station can reconstruct a high-fidelity preview rapidly, enabling practical operational decisions—such as prioritizing scenes for immediate download—and supporting downstream remote-sensing workflows, e.g., vegetation monitoring~\cite{dwyer2000global, ma2008detecting}, urban change detection~\cite{basavaraju2024recent}, and few-shot segmentation~\cite{immanuel2025tackling} without waiting for full data transfer.

Overall, ELMZip provides an effective trade-off between reconstruction quality and transmission cost, offering a deployment-oriented neural compression strategy for resource-constrained Earth observation platforms. The main contributions of this work are summarized as:
\begin{itemize}
\item We introduce ELMZip, an ELM-based neural compression method that enables fast onboard fitting via a convex least-squares formulation without backpropagation.
\item We present an asymmetric downlink protocol that transmits only compact output parameters for efficient reconstruction and preview generation at the ground station.

\item To assess general performance, we evaluate our proposed method on Level-0 and Level-1C datasets covering diverse natural environments.

\end{itemize}

\section{Background and Related Work}
\subsection{Implicit Neural Representations for Images}
INRs fundamentally alter how data is stored and processed~\cite{strumpler2022inrcompress, jo2025pdefuncta, cho2025neural}.
Instead of storing a discrete grid of pixel intensities $V \in \mathbb{R}^{H \times W \times C}$, an INR learns a continuous mapping function $\Phi: \mathbb{R}^2 \rightarrow \mathbb{R}^C$ that maps 2D spatial coordinates $\mathcal{X} = (x, y)$ to their corresponding spectral intensity values $V(\mathcal{X})$. A standard Multi-layer Perceptron (MLP) acts as the function approximator, defined recursively as follows.
\begin{equation}
\Phi(\mathcal{X};\Theta) = W_L(\sigma_{L-1}(\dots \sigma_1(W_1 \cdot \mathcal{X} + b_1)\dots)) + {b}_L,
\end{equation}
where $W_l$ and $b_l$ represent the weights and biases of the $l$-th layer, $L$ is the number of layers, and $\sigma$ is a nonlinear activation function. While standard ReLU networks fail to capture high-frequency image details~\cite{tancik2020fourier}, recent advancements have introduced periodic activations to overcome this spectral bias~\cite{sitzmann2020implicit, liu2024finer, yeom2024fast}. For instance, SIREN~\cite{sitzmann2020implicit} employs a sine activation function, allowing the network to model fine details and their derivatives effectively. Similarly, WIRE~\cite{saragadam2023wire} utilizes a complex Gabor wavelet activation, $e^{i\omega_0 \mathcal{X}} \cdot e^{-|\mathcal{X}|^2}$, to provide optimally concentrated space-frequency localization. Despite their reconstruction quality, these architectures require finding the optimal parameters $\Theta$ by minimizing a non-convex loss function $\mathcal{L} = \sum ||\Phi(\mathcal{X};\Theta) - V(\mathcal{X})||^2$ via gradient descent. This iterative optimization is prohibitively slow for onboard processing, often taking minutes to hours to converge, which is incompatible with the real-time constraints of Low Earth Orbit (LEO) operations.

\subsection{Extreme Learning Machines}
To circumvent the latency of gradient-based training, we look to ELM. The ELM theory posits that for a feedforward network, the input weights and biases $\theta_{\text{fix}}=\{W_{\text{fix}}, b_{\text{fix}}\}$ need not be tuned iteratively. Instead, they can be randomly assigned from a continuous probability distribution and fixed. The training process is thus reduced to learning only the output weights $\mu$. The ELM output is expressed as follows. 
\begin{equation}
\Phi(\mathcal{X};\Theta) = \mathbf{H} \cdot \mu, \quad \mathbf{H} = \sigma(W_{\text{fix}} \cdot \mathcal{X} + b_{\text{fix}})\end{equation}
This formulation allows the optimal output weights $\mu$ to be computed analytically using the output matrix $\mathbf{H}$.

% {\color{blue}
% However, a single global ELM struggles to capture the variations of a high-resolution satellite scene due to global spectral bias~\cite{rahaman2019spectral}. To address this, we employ the domain decomposition strategy found in~\cite{moseley2023finite, anderson2024elm}. In previous research \textbf{(cite)}, the model decomposes the global domain $\Omega$ into overlapping subdomains. The global solution is constructed by summing the weighted contributions of local networks: $u(\mathcal{X}) = \sum_{k=1}^{K} \omega^k(\mathcal{X}) u^k(\mathcal{X})$, where $\omega^k(\mathcal{X})$ are smooth window functions forming a partition of unity such that $\sum_{k=1}^{K}\omega^k(\mathcal{X}) = 1$. This approach localizes the complexity, allowing multiple small, fast ELMs to model intricate local features accurately without the computational burden of a massive global network.
% }
% However, a single global ELM struggles to capture the variations of a high-resolution satellite scene due to global spectral bias~\cite{rahaman2019spectral}. To address this, we employ the domain decomposition strategy found in~\cite{moseley2023finite, anderson2024elm}. In previous research \textbf{(cite)}, the model decomposes the global domain $\Omega$ into overlapping subdomains. The global solution is constructed by summing the weighted contributions of local networks. This approach localizes the complexity, allowing multiple small, fast ELMs to model intricate local features accurately without the computational burden of a massive global network.

\section{Proposed Method: ELMZip}
We propose the ELMZip framework, which adapts the ELM with domain decomposition frameworks to the problem of high-speed multispectral image compression.

The core objective is to approximate the multispectral image function $V(\mathcal{X})$ using a collection of localized Extreme Learning Machines, enabling highly efficient data transmission. Our ELMZip framework can be summarized as follows:
\begin{itemize}
    \item \textbf{Onboard Encoding:} The satellite generates the random input layer using a fixed initialization, computes the analytical solution for the output weights $\mu$.

    \item \textbf{Transmission:} The downlink payload consists only of the quantized output weights $\mu$. The input weights and biases $\theta_{\text{fix}}=\{W_{\text{fix}}, b_{\text{fix}}\}$ are never transmitted.

    \item \textbf{Ground Decoding:} The ground station receives $\mu$ from onboard satellite and combines it with the shared fixed initialization $\theta_{\text{fix}}$ to reconstruct the high-dimensional multispectral satellite image.
\end{itemize}

\begin{table}[t!]
\small
\centering
\caption{Geographic Coordinates of Benchmark Dataset}
\renewcommand{\arraystretch}{0.8}
\resizebox{\columnwidth}{!}{%
\begin{tabular}{lcccccc}
\specialrule{1pt}{2pt}{2pt}
\textbf{Data} & \textbf{Level} &\textbf{Latitude} & \textbf{Longitude} & \textbf{Environments} \\
% \cmidrule(lr){1-4}
\specialrule{1pt}{2pt}{2pt}
\textbf{Antuco}  & L0  & 37$^{\circ}$35'05"S & 71$^{\circ}$13'37"W & Volcano \\
\textbf{Puszta}  & L0  & 47$^{\circ}$28'18"N & 19$^{\circ}$53'27"E & Grassland \\
\textbf{Andaman}  & L0  & 12$^{\circ}$26'54"N & 93$^{\circ}$56'27"E & Marine \\
\textbf{Cairo}  & L1C  & 30$^{\circ}$01'29"N & 31$^{\circ}$55'18"E & Desert \\
\textbf{Merapi} & L1C  & 07$^{\circ}$32'29"S & 110$^{\circ}$26'46"E & Volcano \\
\textbf{Seoul}  & L1C  & 37$^{\circ}$31'30"N & 126$^{\circ}$55'36"E & Urban \\
\specialrule{1pt}{2pt}{2pt}
\end{tabular}}\label{tbl:dataset_coord}
\end{table}

\begin{figure}[t!]
\centering
\vspace{-0.3cm}
\subfloat[Antuco (L0)]{\includegraphics[width=0.30\columnwidth]{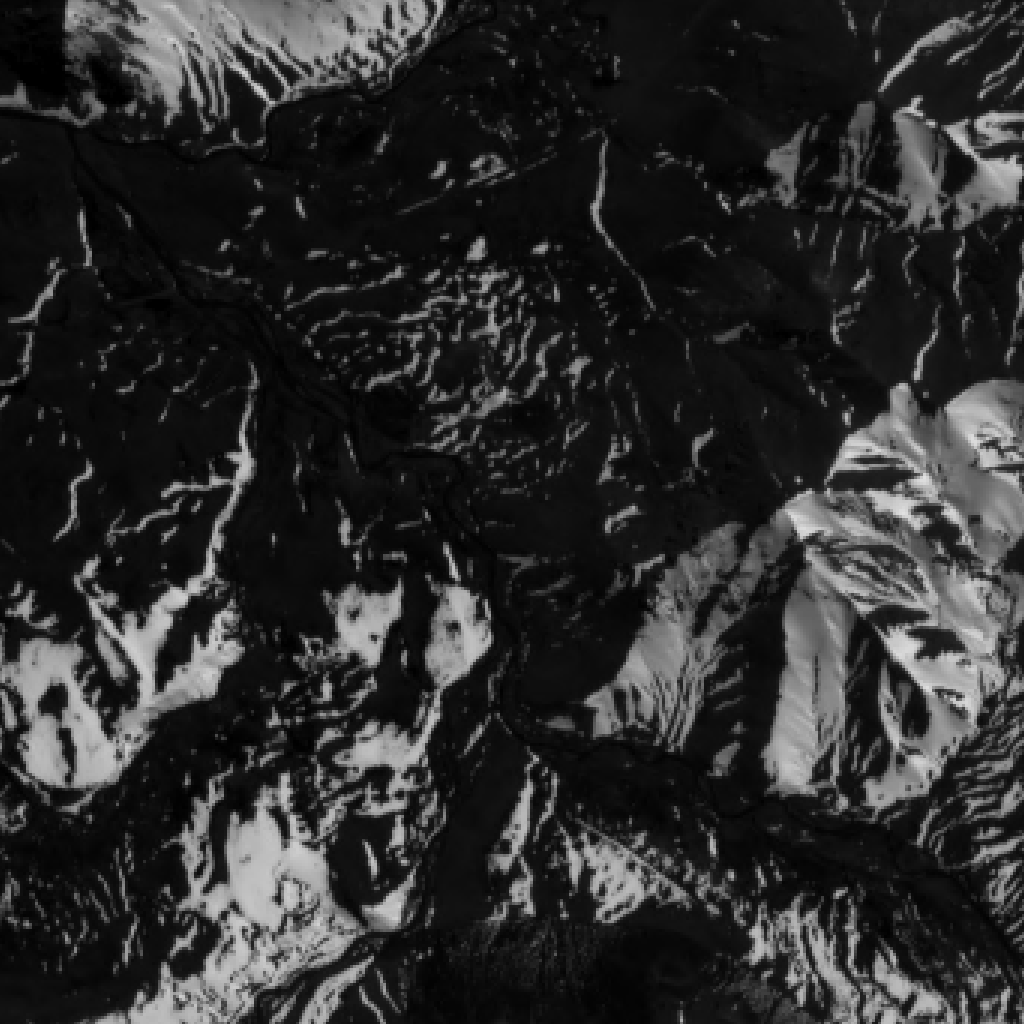}}\hfill
\subfloat[Puszta (L0)]{\includegraphics[width=0.30\columnwidth]{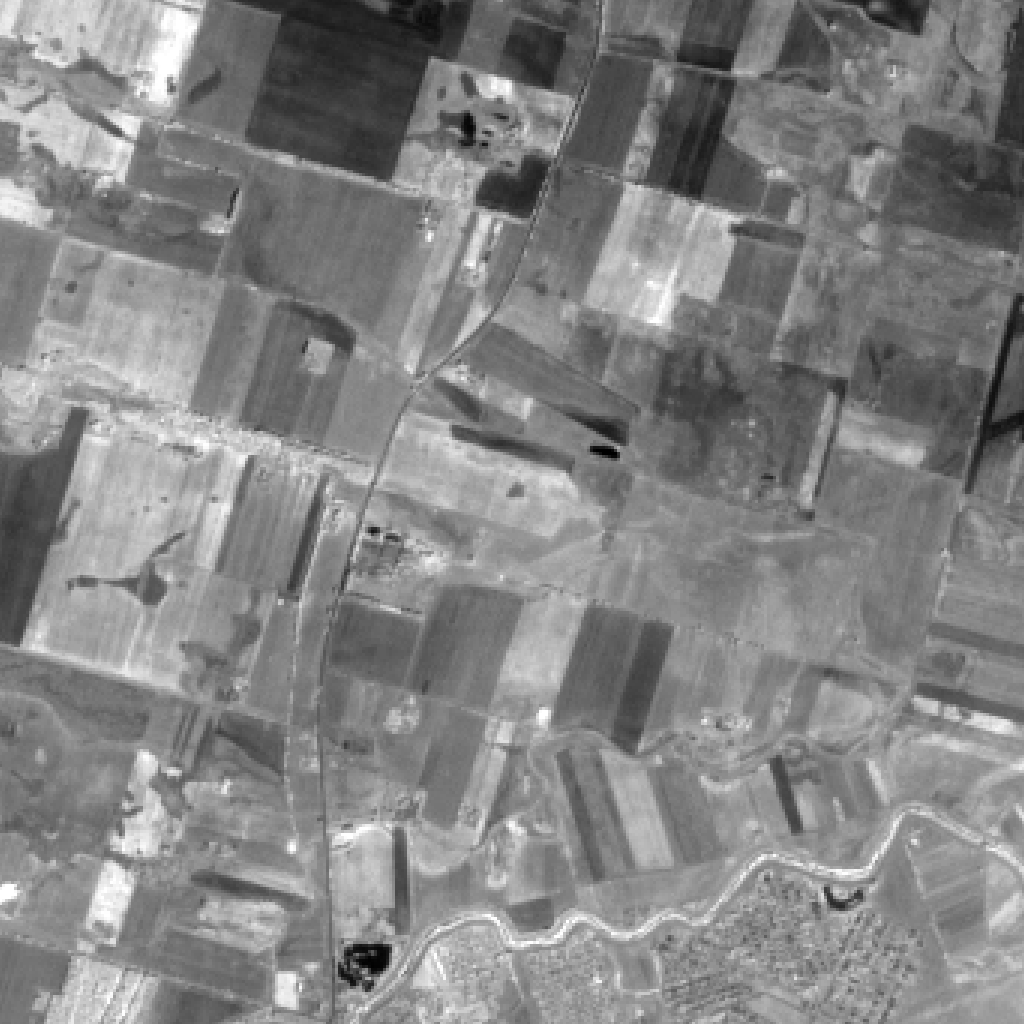}}\hfill
\subfloat[Andaman (L0)]{\includegraphics[width=0.30\columnwidth]{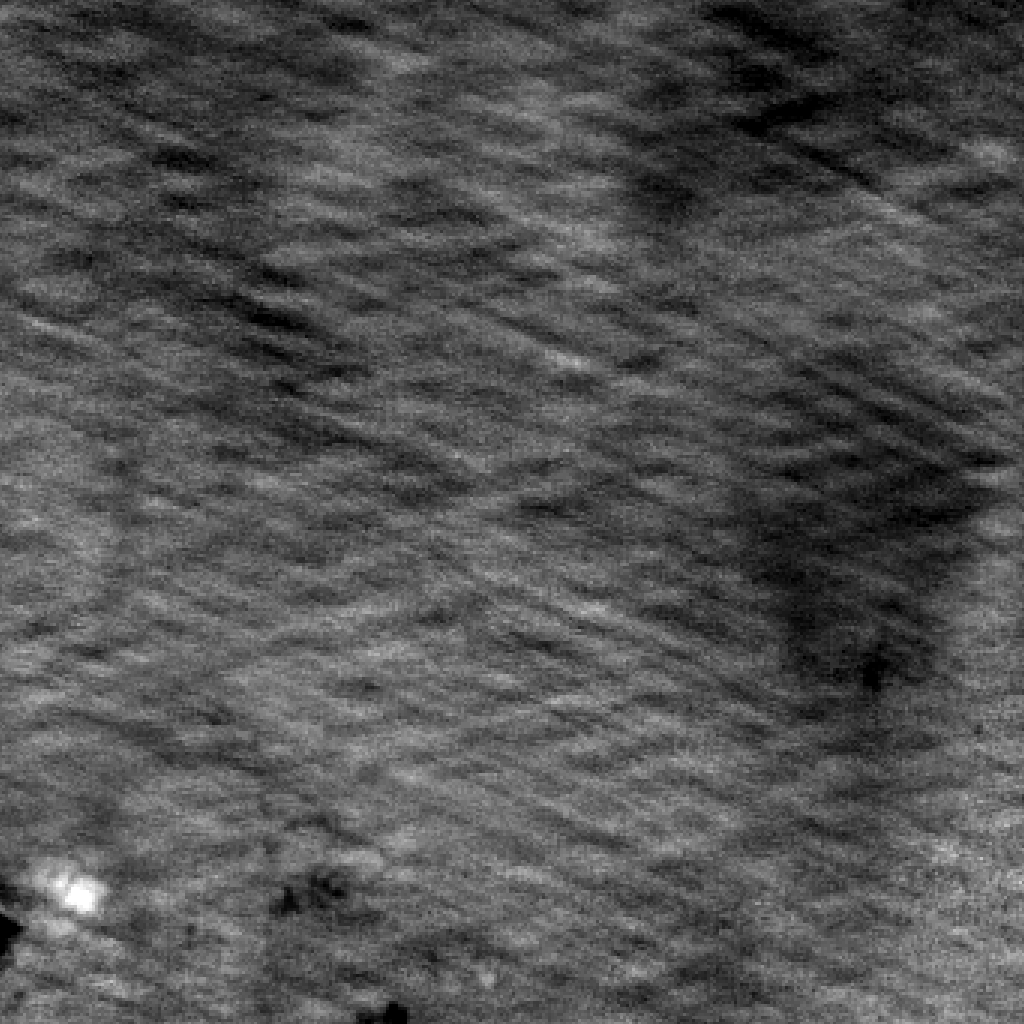}}\\
\subfloat[Cairo (L1C)]{\includegraphics[width=0.30\columnwidth]{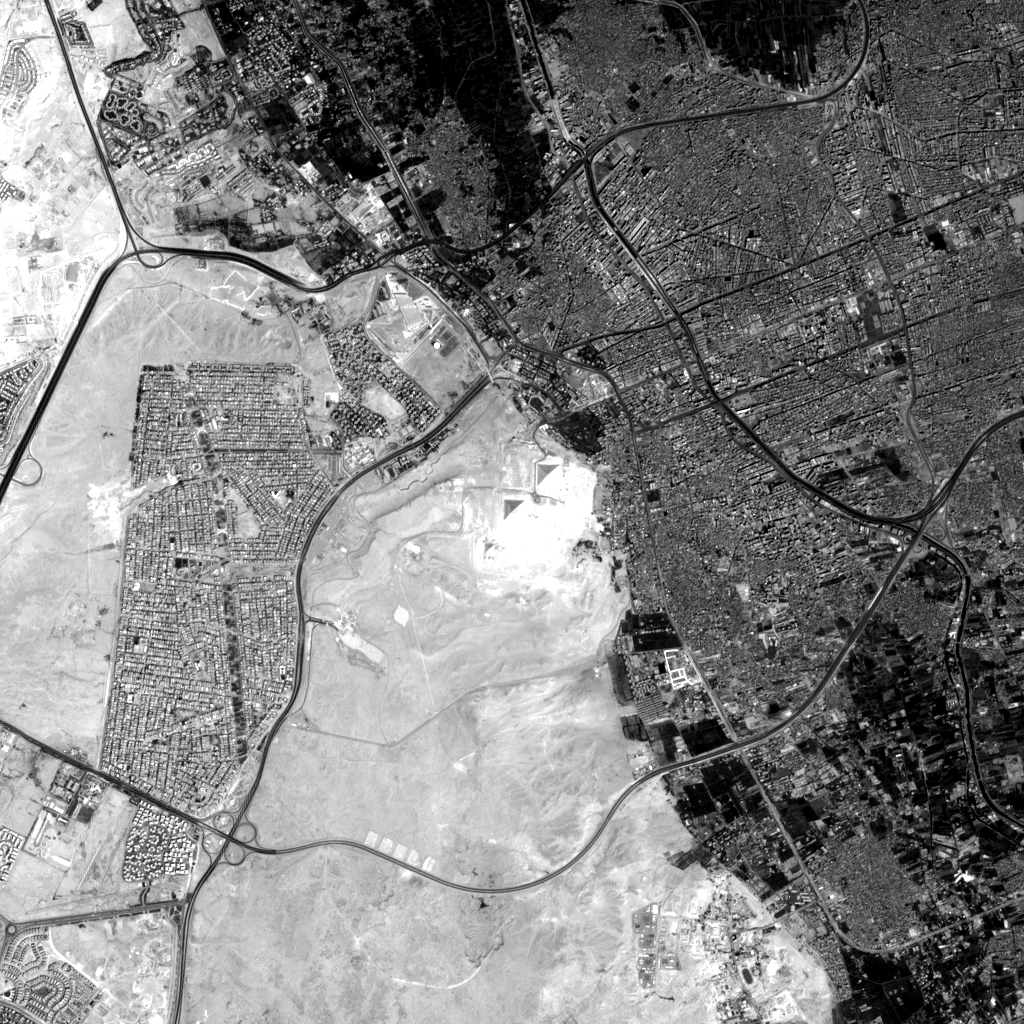}}\hfill
\subfloat[Merapi (L1C)]{\includegraphics[width=0.30\columnwidth]{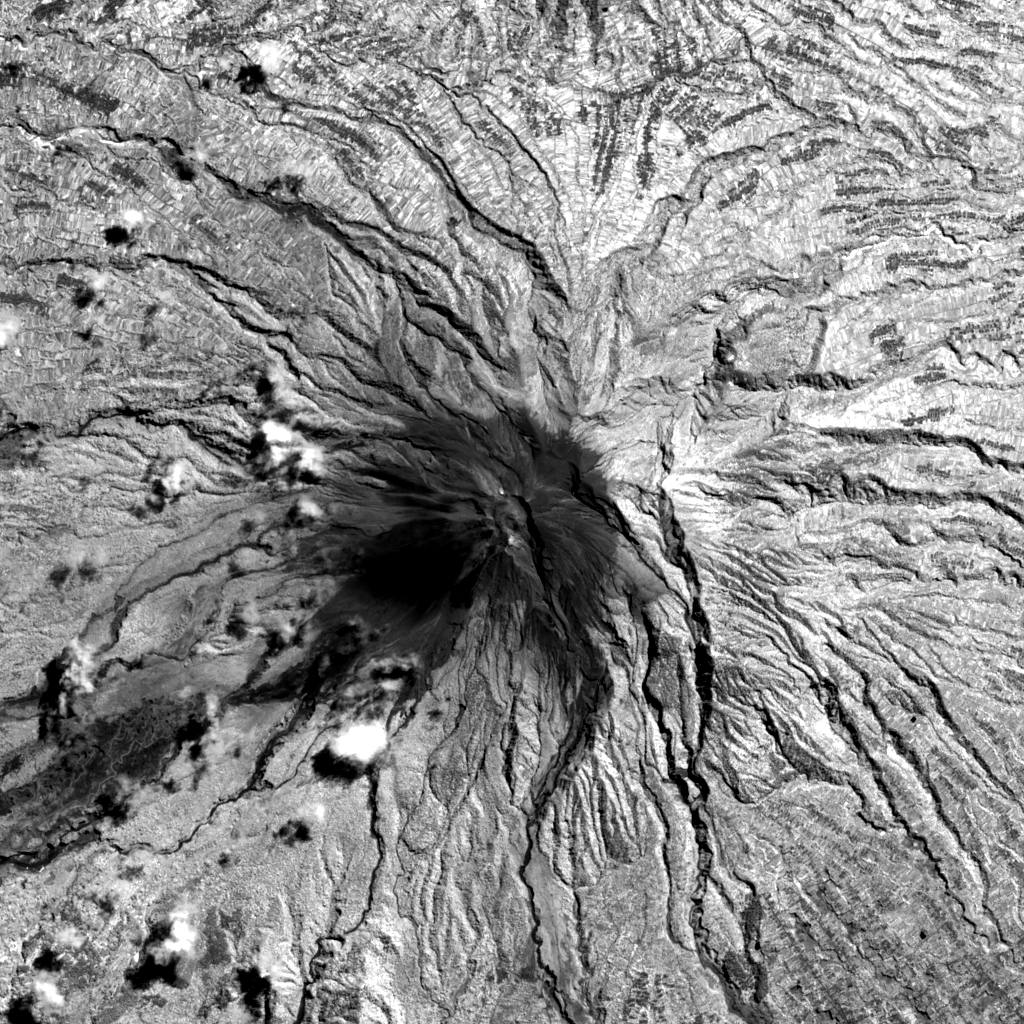}}\hfill
\subfloat[Seoul (L1C)]{\includegraphics[width=0.30\columnwidth]{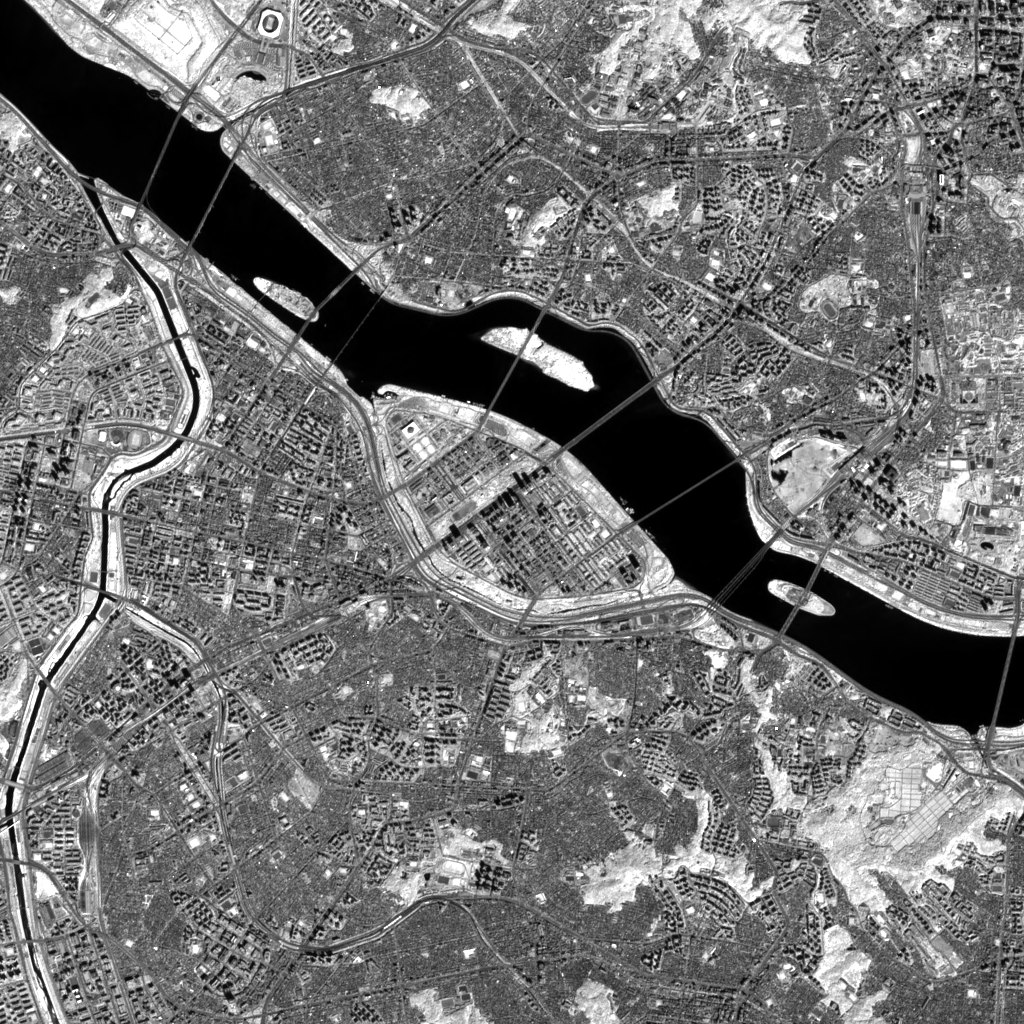}}
\caption{Sentinel-2 benchmark dataset. The dataset consists of six distinct regions covering diverse environments across both Level-0 ((a)-(c)) and Level-1C ((d)-(f)) data products.}
\label{fig:bench_img}
\end{figure}

\subsection{Extreme Learning Machine with Domain Decomposition}

A single global ELM struggles to capture the variations of a high-resolution satellite scene due to global spectral bias~\cite{rahaman2019spectral}. To address this, we employ the domain decomposition strategy found in~\cite{moseley2023finite, anderson2024elm}.

Given a high-resolution multispectral image, we first decompose the spatial domain $\Omega$ into $K$ overlapping subdomains $\{\Omega_k\}_{k=1}^K$. For each subdomain, we assign a local ELM approximator $\phi^k(\mathcal{X}; \Theta^{k})$. 
% \begin{equation}
% \phi^k(\mathcal{X};{{\Theta}^k})= \sum_{j=1}^{J} \mu^j \sigma(W^j \cdot \mathcal{X} + b^j).\label{eq:sub_elm}
% \end{equation}
% In~\eqref{eq:sub_elm}, w
We utilize sine function for the activation function $\sigma$. The global image reconstruction is defined as the weighted sum of these local models:
% \begin{equation}
% \hat{V}(\mathcal{X};{\theta})=\sum_{k=1}^{K} \omega^k(\mathcal{X}) \cdot \phi^k(\mathcal{X};{{\theta}^k}), \quad \Theta^k=\{\theta^k_{\text{fix}}, \mu^k\}
% \end{equation}
\begin{equation}\hat{V}(\mathcal{X};{\Theta})=\sum_{k=1}^{K} \omega^k(\mathcal{X}) \cdot \phi^k(\mathcal{X};{{\Theta}^k}), \quad 
\Theta = \{\theta^k_\text{fix}, \mu^k\}_{k=1}^K
\end{equation}
where $\omega^k(\mathcal{X})$ are smooth window functions forming a partition of unity such that $\sum_{k=1}^{K}\omega^k(\mathcal{X}) = 1$. Concretely, the input parameters $\theta^k_{\text{fix}}=\{{W}_n^{k} , b_n^{k}\}$ are initialized using a built-in function in PyTorch. By sharing the fixed parameters $\theta_\text{fix}=\{\theta_\text{fix}^k\}_{k=1}^K$ between the satellite and the ground station, these parameters become deterministic and do not need to be transmitted. 
% Crucially, the input parameters for each local network, $\theta_{k}^{\text{fix}} = {W_{k}^{\text{fix}}, b_{k}^{\text{fix}}}$, are initialized using a pseudo-random number generator with a fixed seed. By sharing the seed between the satellite and the ground station, these random projection parameters remain deterministic and need not be transmitted.
% \subsection{Analytical Optimization via Linear Least Squares}

\subsection{Optimization for ELMZip}
Unlike standard INRs that minimize loss via backpropagation, ELMZip formulates the training as a linear least-squares problem. As shown in Fig.~\ref{fig:model_archi}, for a specific subdomain $k$, we aim to minimize the reconstruction error between the model output and the ground truth pixel values $V^k_{\text{gt}}$. Since the input projection parameters are fixed, the optimization objective for the output weights $\mu^{k}$ is strictly convex:
% \begin{equation}
% \min_{\mu^{k}} \frac{1}{2} \| \mathbf{H}^k \mu^{k} - {V}^k_{\text{gt}} \|_2^2 + \frac{\lambda}{2} \| \mu^{k} \|_2^2
% \end{equation}
\begin{equation}
\min_{\mu^{k}} \frac{1}{2} \| \mathbf{H}^k \mu^{k} - {V}^k_{\text{gt}} \|_2^2
\end{equation}
In this equation, $\mathbf{H}^k$ is the hidden layer activation matrix computed as $\mathbf{H}^k = \omega^k(\mathcal{X}) \cdot \sigma(W_\text{fix}^{k} \cdot \mathcal{X} + b_\text{fix}^{k})$. The optimal solution $\hat{\mu}^{k}$ is obtained in a single computational step by solving the normal equations:

\begin{equation}
\hat{\mu}^{k} = \left( \mathbf{H}^{k\top} \mathbf{H}^k \right)^{-1} \mathbf{H}^{k\top} V^k_{gt}
\end{equation}
This closed-form solution reduces training to matrix multiplications and a linear solve, which can be efficiently accelerated by CUDA-optimized linear algebra libraries on embedded GPUs such as the NVIDIA Jetson Nano. As a result, the model can fit each image patch rapidly.
\begin{table*}[t!]
\caption{PSNR($\uparrow$) and SSIM($\uparrow$) comparison for the ELMZip (ours) against baseline methods}
\renewcommand{\arraystretch}{0.8}
\resizebox{\textwidth}{!}{%
\begin{tabular}{lcccccccccccc}
\specialrule{1pt}{2pt}{2pt}
\multirow{3}{*}{Model} & \multicolumn{2}{c}{Antuco (L0)} & \multicolumn{2}{c}{Puszta (L0)} & \multicolumn{2}{c}{Andaman (L0)} & \multicolumn{2}{c}{Cairo (L1C)} & \multicolumn{2}{c}{Merapi (L1C)} & \multicolumn{2}{c}{Seoul (L1C)}
\\ \cmidrule(lr){2-3}\cmidrule(lr){4-5}\cmidrule(lr){6-7}\cmidrule(lr){8-9}\cmidrule(lr){10-11}\cmidrule(lr){12-13}
                       & PSNR   & SSIM   & PSNR   & SSIM   & PSNR   & SSIM   & PSNR   & SSIM   & PSNR   & SSIM   & PSNR   & SSIM  \\
\midrule
MLP
& 14.521 & 0.475
& 17.979 & 0.466
& 21.113 & 0.620
& 15.433 & 0.772
& 13.689 & 0.410
& 12.877 & 0.211 \\

SIREN
& 22.302 & 0.940
& 23.970 & 0.906
& 24.388 & 0.854
& 17.842 & 0.883
& 15.852 & 0.715
& 16.026 & 0.735 \\

FFN
& 22.818 & 0.948
& 22.177 & 0.825
& 24.704 & 0.856
& 17.992 & 0.884
& 15.769 & 0.671
& 16.097 & 0.720 \\

GaussNet
& 16.741 & 0.744
& 21.002 & 0.793
& 22.535 & 0.764
& 16.843 & 0.846
& 15.016 & 0.631
& 14.955 & 0.624 \\

WIRE
& 16.469 & 0.725
& 19.558 & 0.721
& 21.759 & 0.716
& 15.474 & 0.787
& 14.705 & 0.608
& 14.461 & 0.582 \\

ELMZip
& \textbf{26.943} & \textbf{0.981}
& \textbf{32.426} & \textbf{0.988}
& \textbf{30.039} & \textbf{0.965}
& \textbf{20.961} & \textbf{0.946}
& \textbf{20.847} & \textbf{0.925}
& \textbf{19.585} & \textbf{0.899} \\
\specialrule{1pt}{2pt}{2pt}
\end{tabular}}
\label{tbl:main_results}
\end{table*}
\begin{figure*}[t!]
\centering
% \vspace{-0.1cm}
% ===================== Antuco (L0, B8) =====================
\subfloat[Ground Truth]{\includegraphics[width=0.275\columnwidth]{Figures/L0_antuco.png}}\hfill
\subfloat[MLP]{\includegraphics[width=0.275\columnwidth]{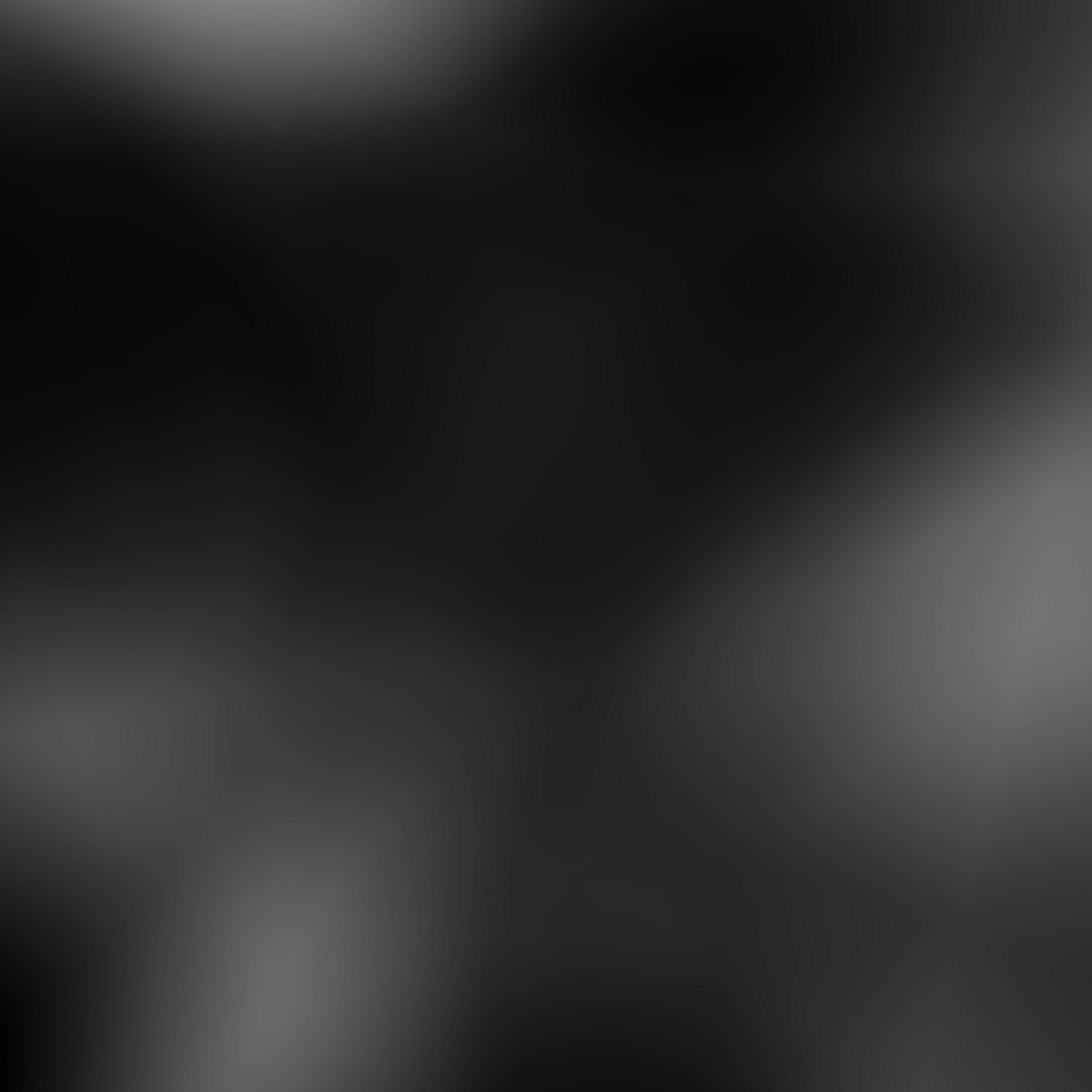}}\hfill
\subfloat[SIREN]{\includegraphics[width=0.275\columnwidth]{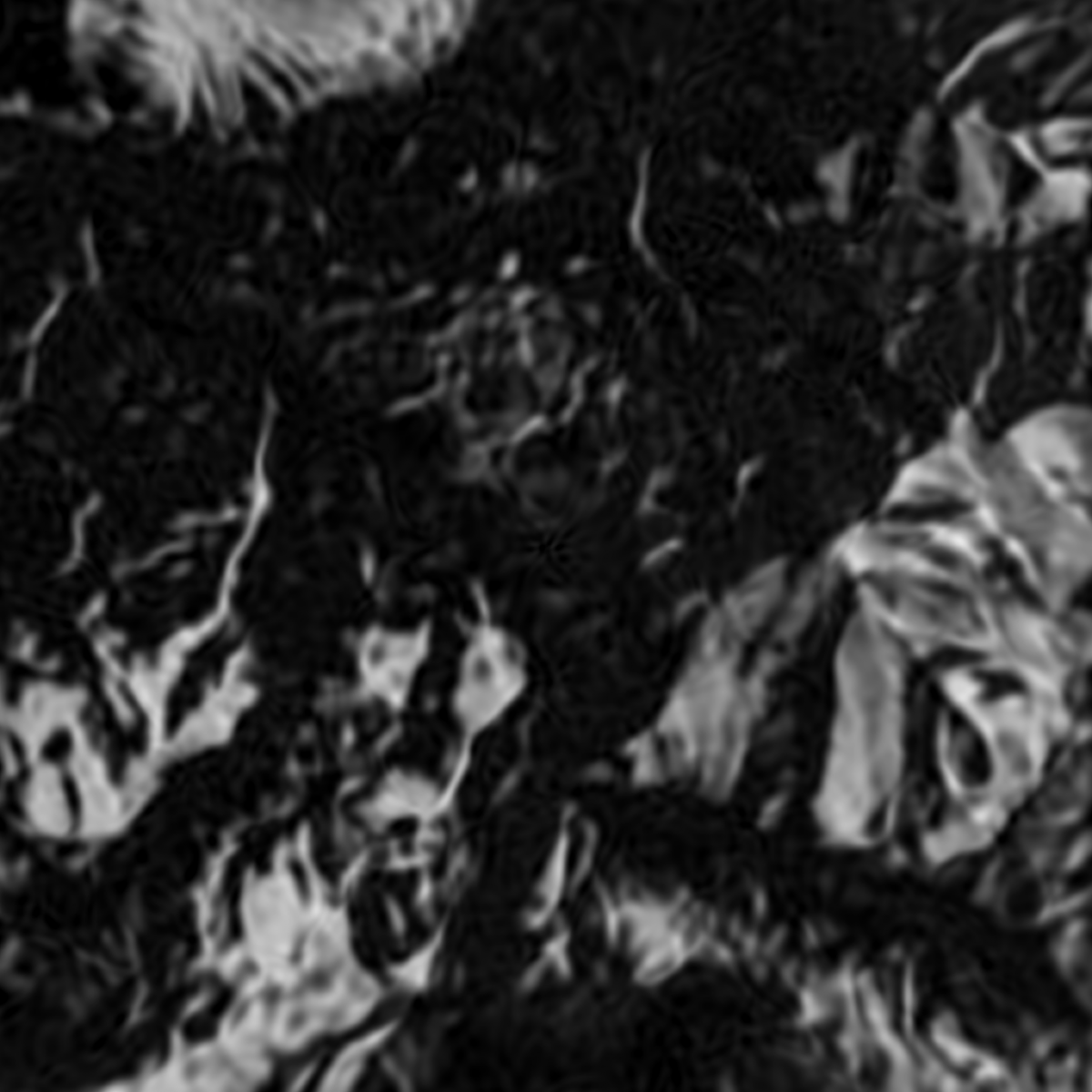}}\hfill
\subfloat[FFN]{\includegraphics[width=0.275\columnwidth]{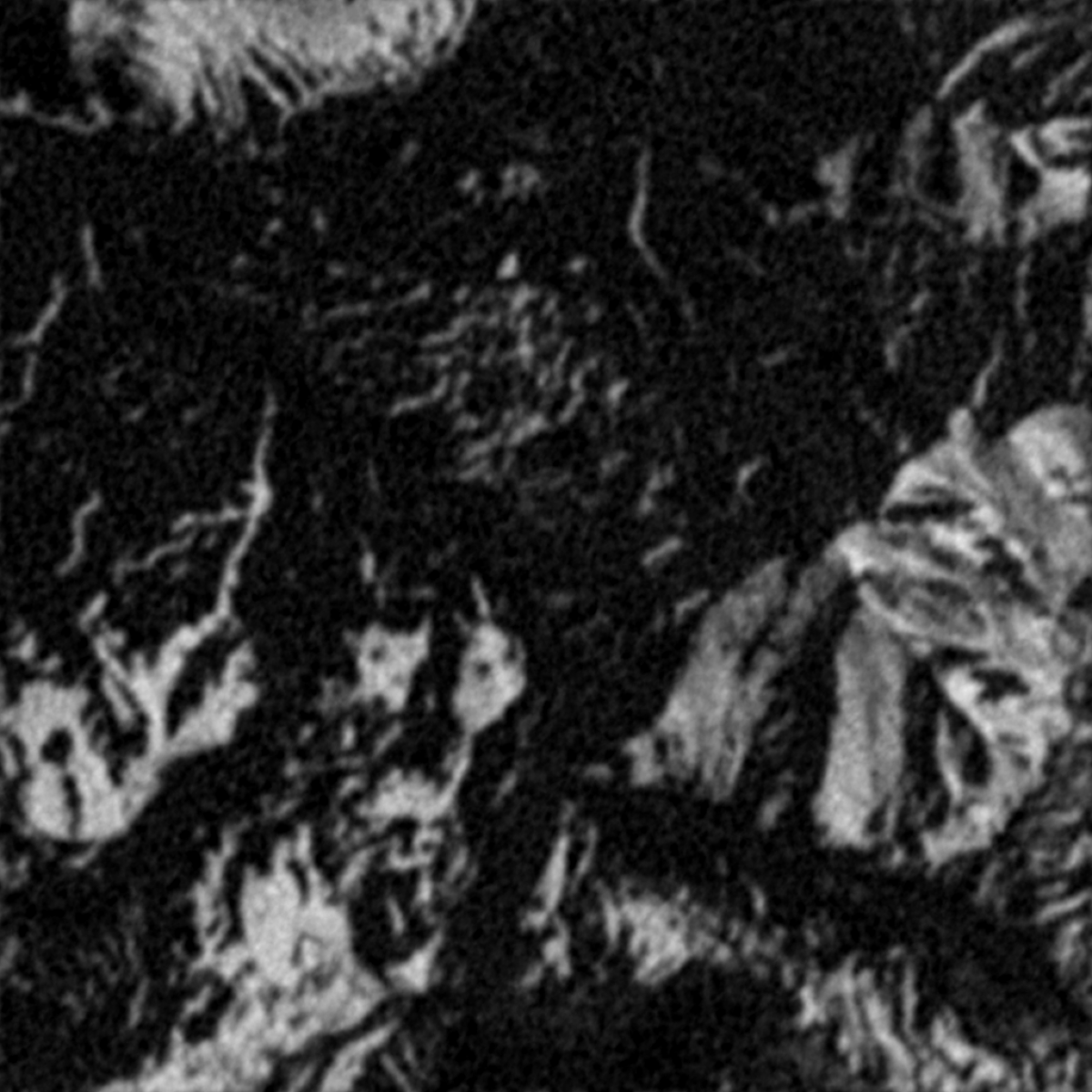}}\hfill
\subfloat[GaussNet]{\includegraphics[width=0.275\columnwidth]{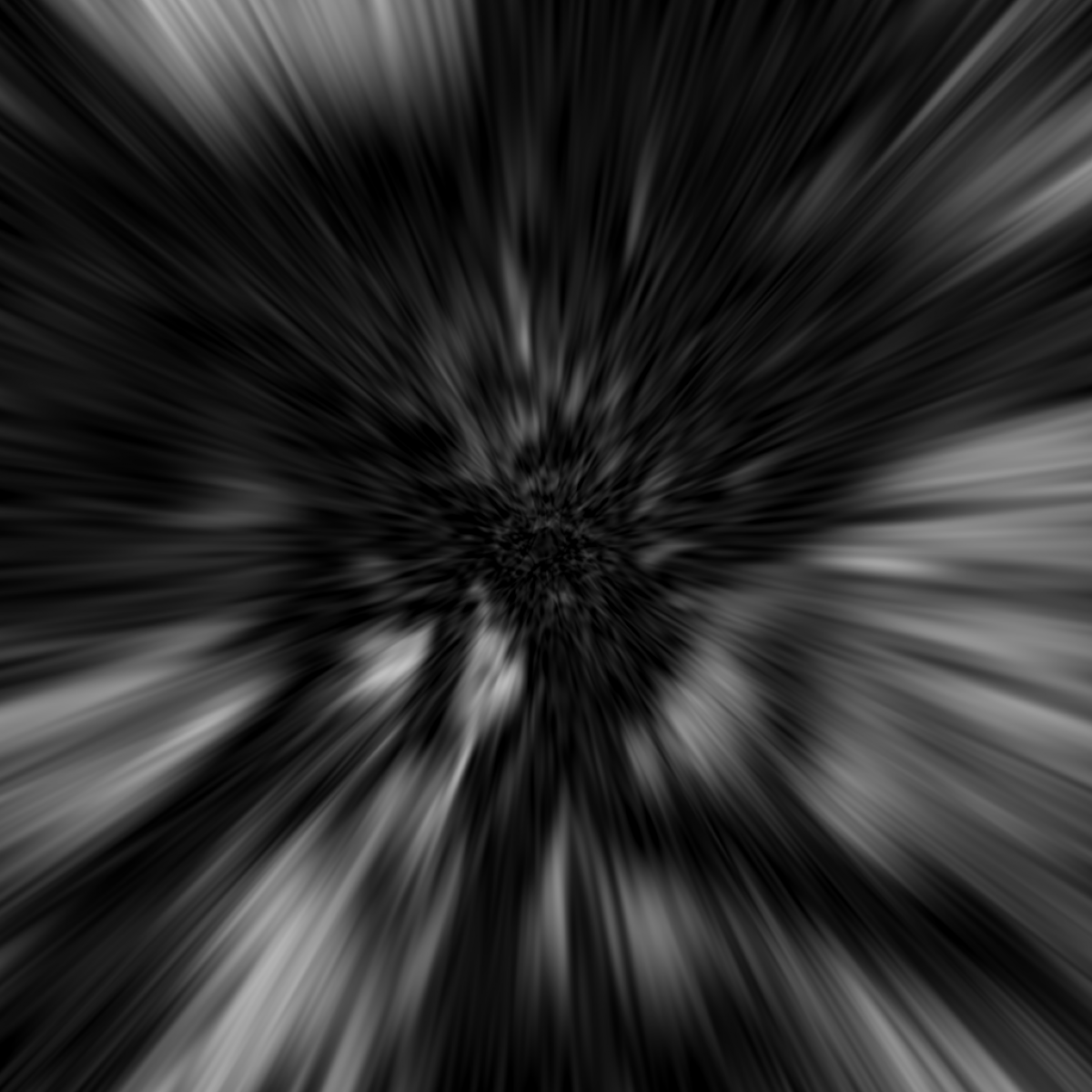}}\hfill
\subfloat[WIRE]{\includegraphics[width=0.275\columnwidth]{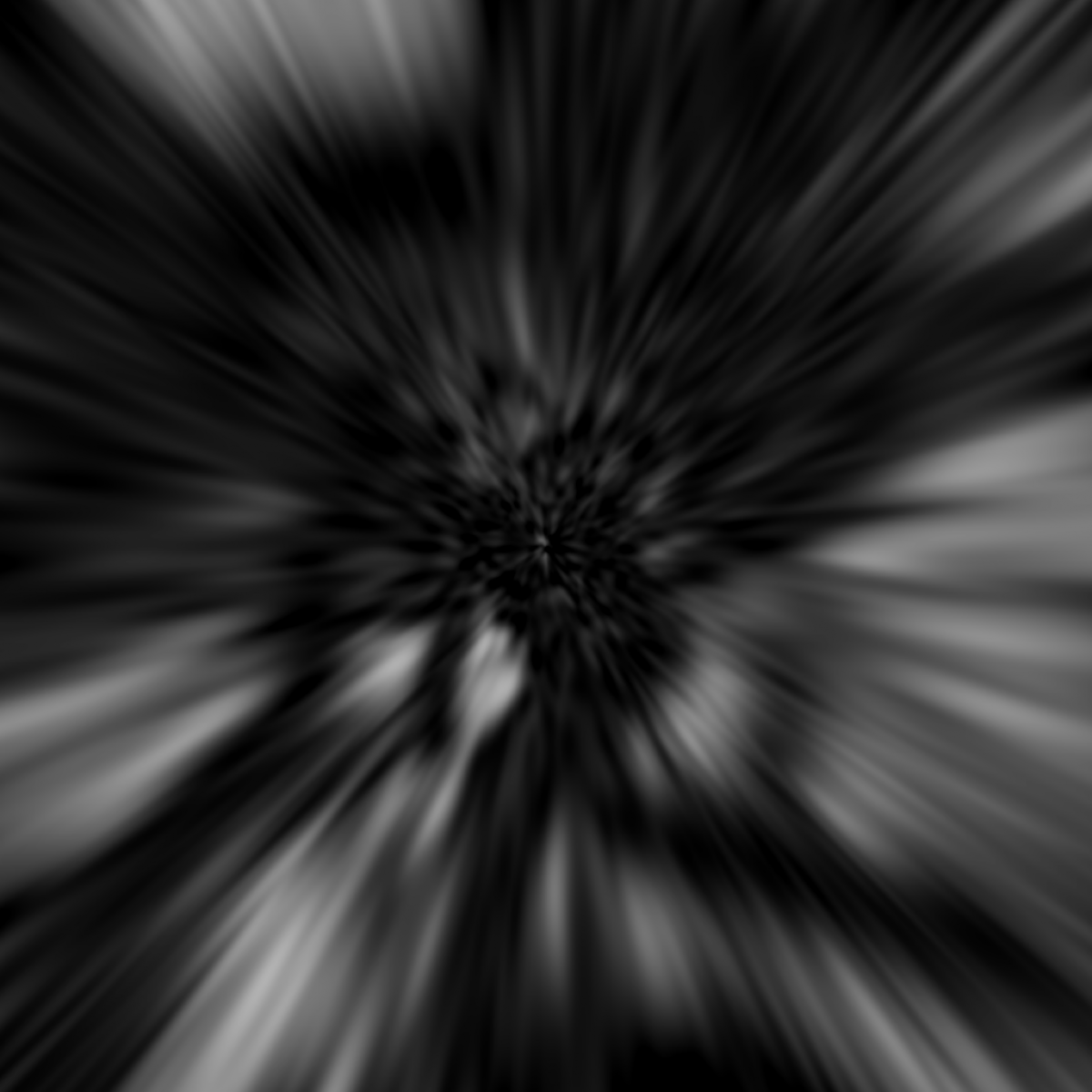}}\hfill
\subfloat[ELMZip]{\includegraphics[width=0.275\columnwidth]{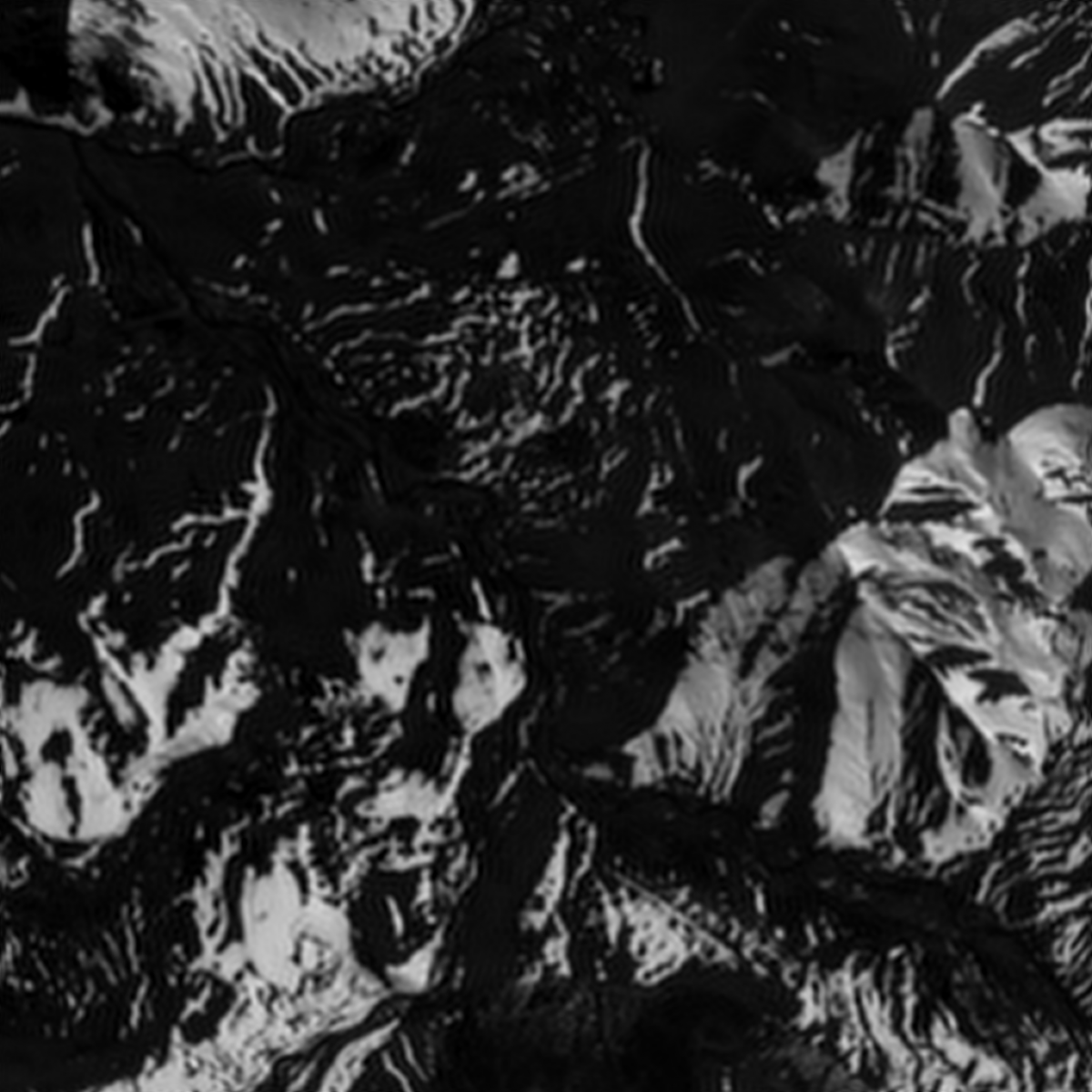}}\\

% ===================== Cairo (L1C, B3) =====================
\subfloat[Ground Truth]{\includegraphics[width=0.275\columnwidth]{Figures/L1_cairo_band03.png}}\hfill
\subfloat[MLP]{\includegraphics[width=0.275\columnwidth]{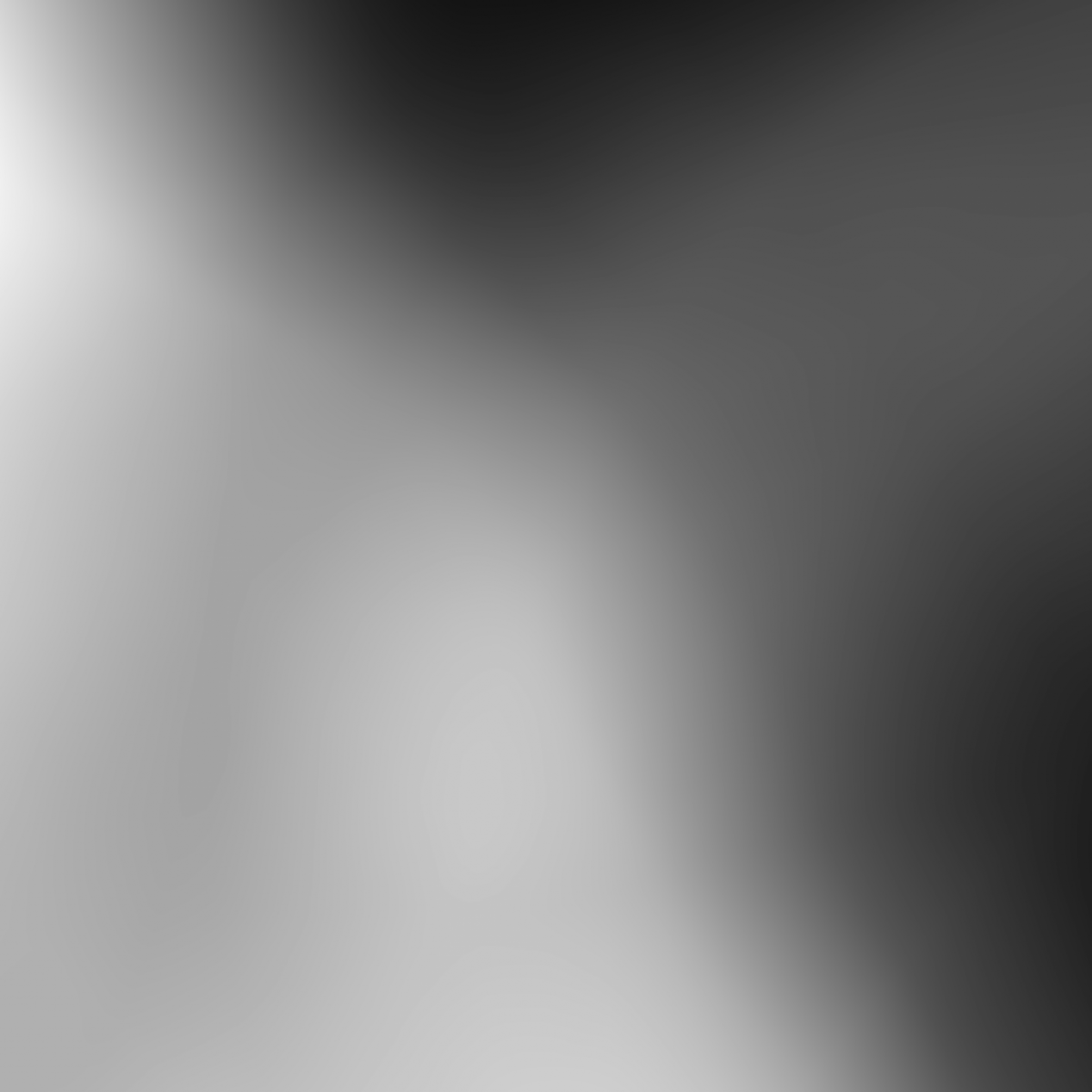}}\hfill
\subfloat[SIREN]{\includegraphics[width=0.275\columnwidth]{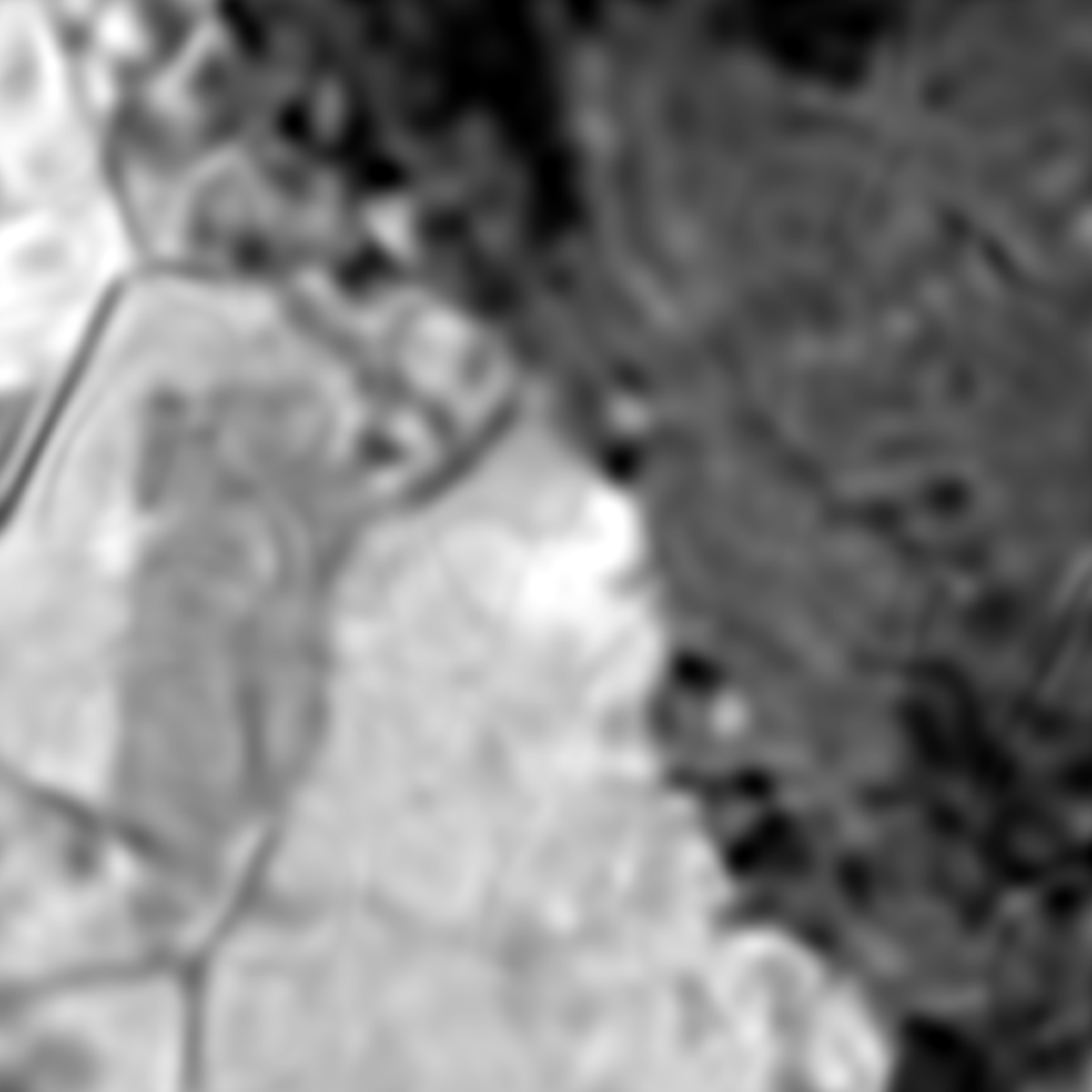}}\hfill
\subfloat[FFN]{\includegraphics[width=0.275\columnwidth]{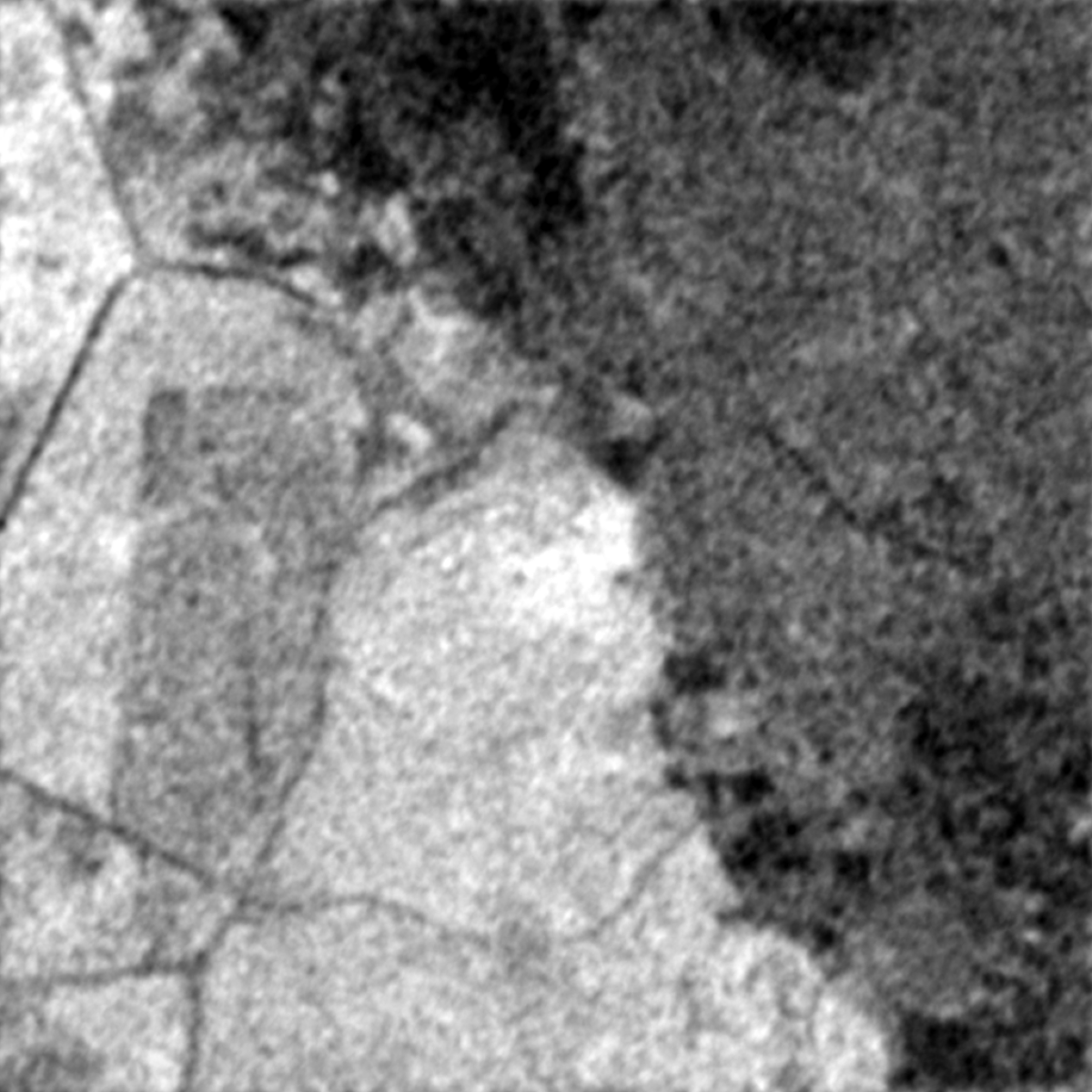}}\hfill
\subfloat[GaussNet]{\includegraphics[width=0.275\columnwidth]{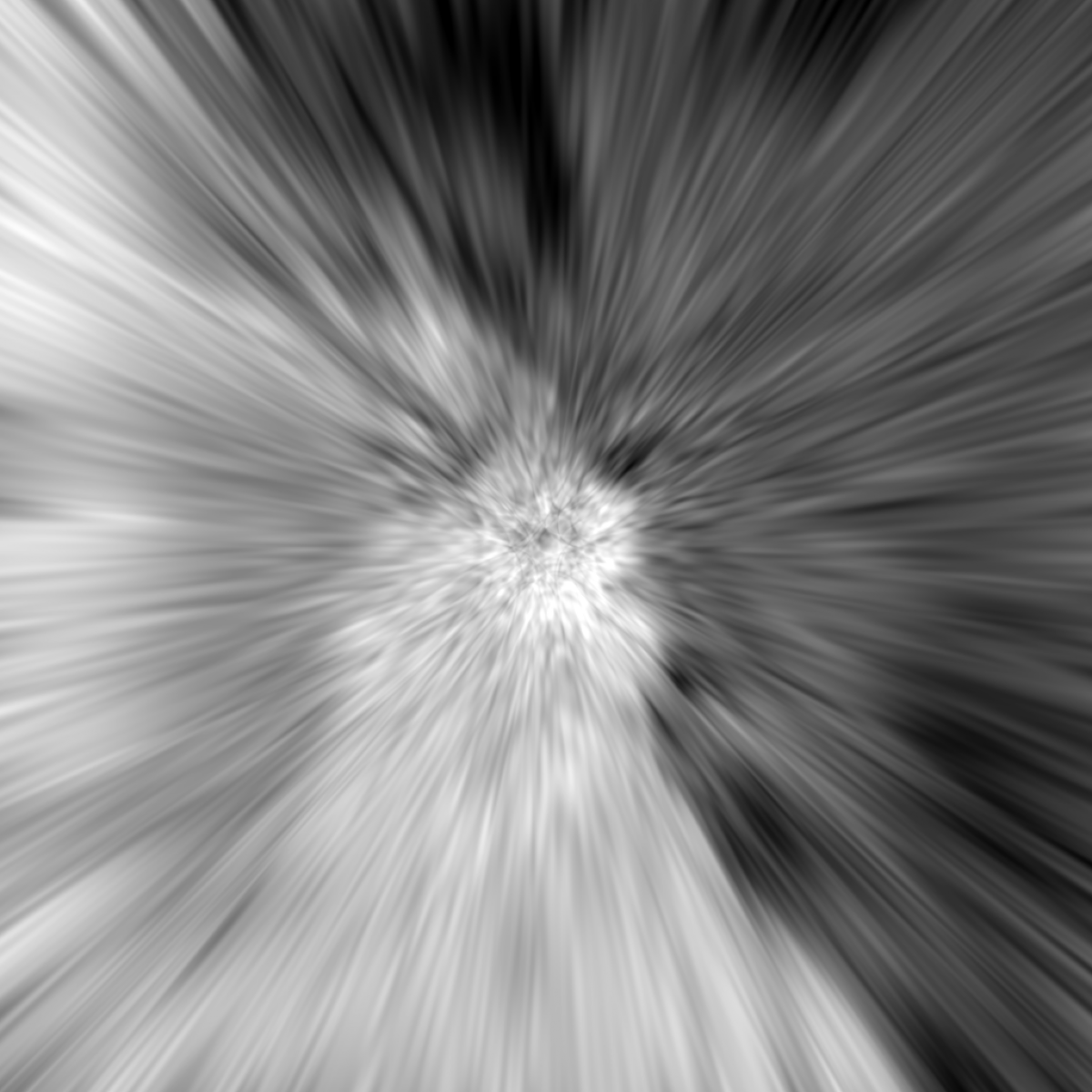}}\hfill
\subfloat[WIRE]{\includegraphics[width=0.275\columnwidth]{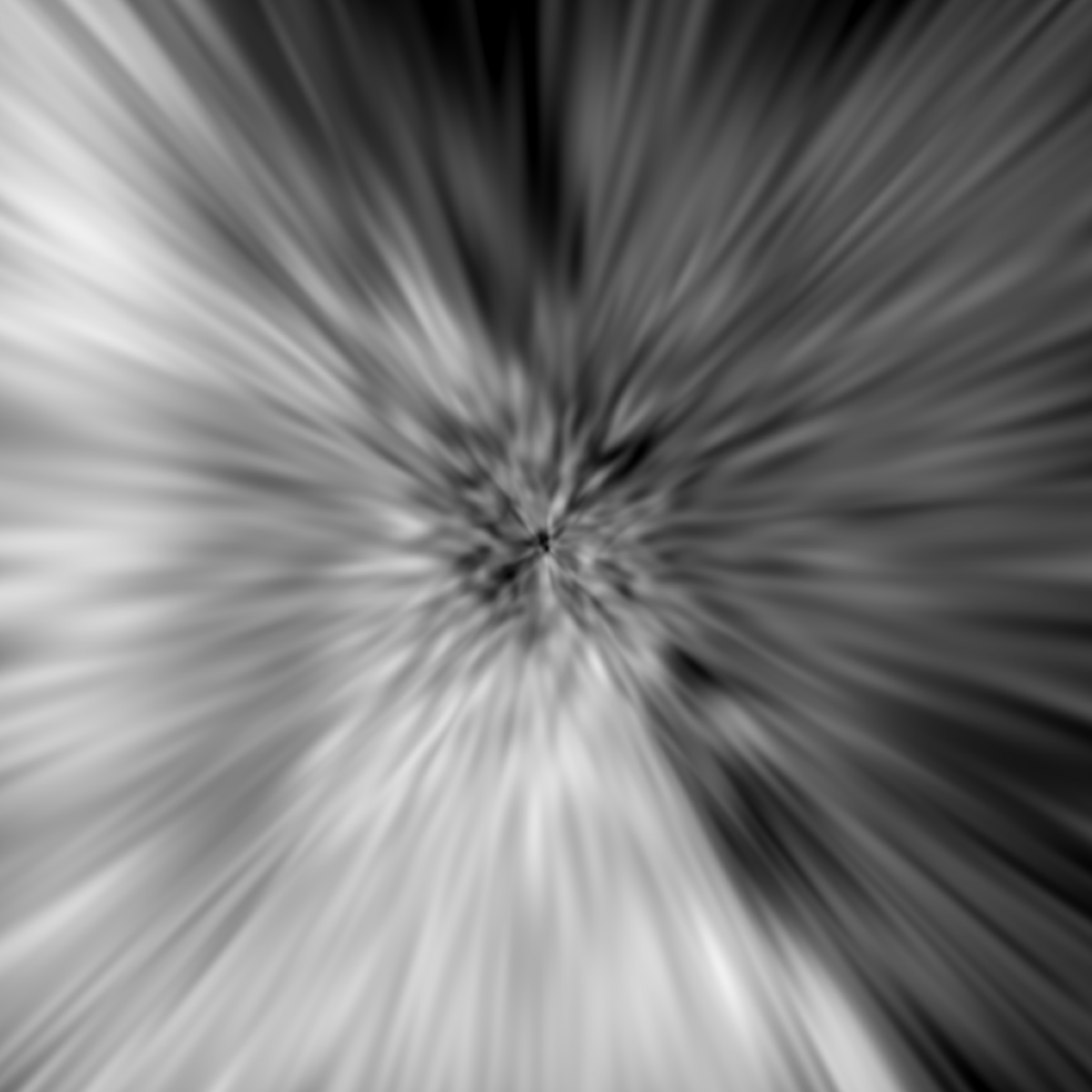}}\hfill
\subfloat[ELMZip]{\includegraphics[width=0.275\columnwidth]{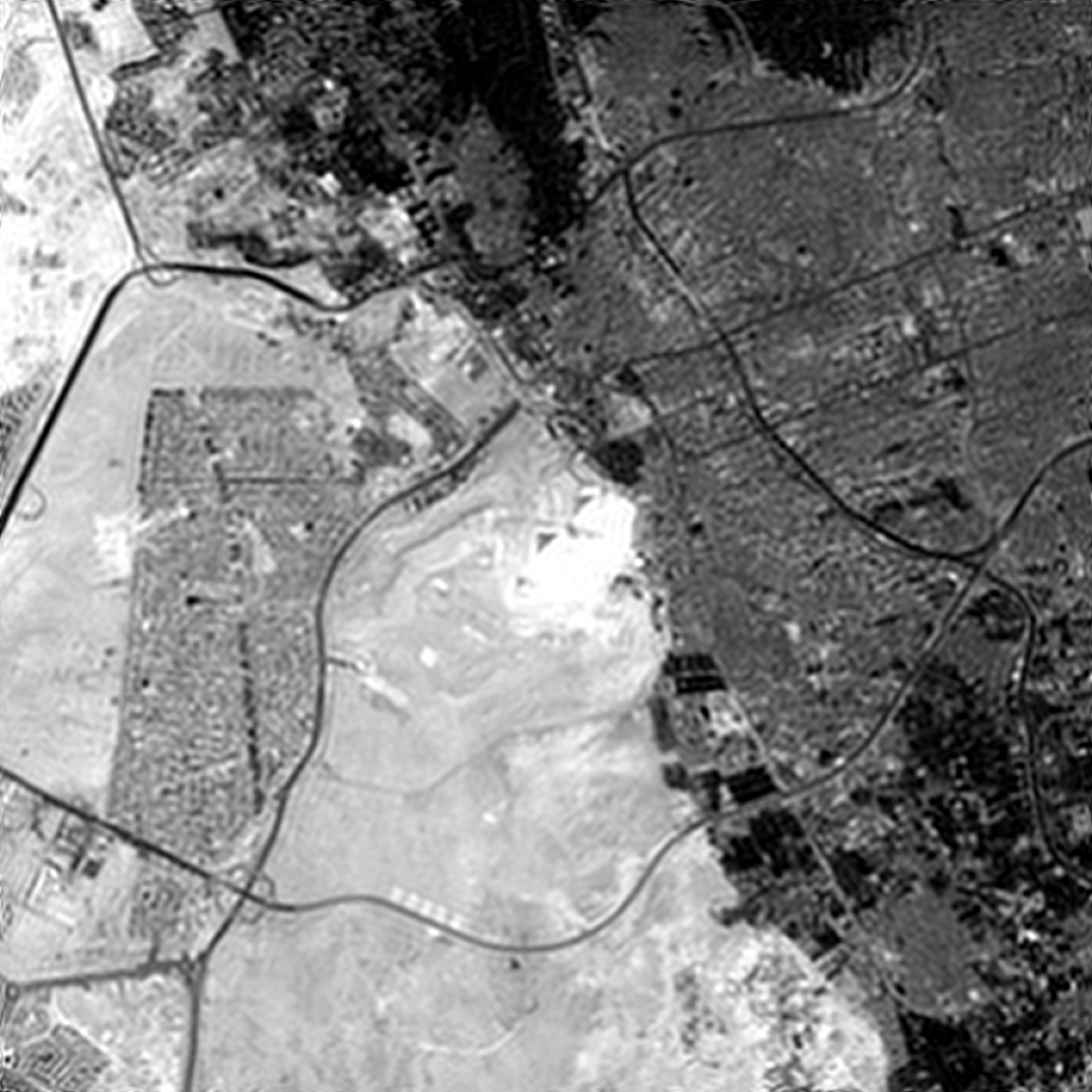}}\\

\caption{\textbf{Qualitative reconstruction comparisons on Sentinel-2 MSI under unequal energy budgets.}
The top row shows Antuco (L0, B8) and the bottom row shows Cairo (L1C, B3).
INR baselines are optimized with approximately 10$\times$ higher electrical energy than ELMZip,
yet ELMZip produces reconstructions that are visually closest to the ground truth.}
\label{fig:main_results}
\end{figure*}

\section{Experiments}
To validate the proposed method under realistic onboard constraints, we utilized Sentinel-2 MSI Level-0 and Level-1C data collected from six distinct geographic regions representing diverse land covers\cite{meoni2024unlocking}, as summarized in Table~\ref{tbl:dataset_coord} and visualized in Fig.~\ref{fig:bench_img}. In particular, to approximate an onboard satellite scenario as closely as possible, our benchmark suite explicitly includes Level-0 (L0) products, which are the most similar to raw sensor measurements and therefore capture the practical characteristics of minimally processed imagery available in early downlink pipelines. The onboard processing environment was simulated using an NVIDIA Jetson Nano, a representative low-power edge computing platform for nanosatellites. 

\subsection{Experimental Setup}
We compared ELMZip against representative INR baselines, including an MLP with ReLU, SIREN~\cite{sitzmann2020implicit}, Fourier Feature Networks (FFN)~\cite{tancik2020fourier}, GaussNet~\cite{ramasinghe2022beyond}, and WIRE~\cite{saragadam2023wire}. All methods were implemented in PyTorch. For a controlled and fair comparison, all benchmark images were standardized to a fixed spatial dimension of $1024$ so that the evaluation is not confounded by varying scene sizes or different coordinate sampling densities. We also designed the evaluation to reflect practical downlink constraints by enforcing neural compression to achieve at least an approximately $10\times$ reduction in payload relative to the uncompressed image. This was implemented by constraining the transmitted representation size across methods, for example through a common parameter budget and quantized payload settings.
We measure the electrical energy consumed during fitting. Due to iterative backpropagation, INR baselines require approximately $10\times$ more energy than ELMZip to reach their reported performance. We evaluated performance using two key metrics: Peak Signal-to-Noise Ratio (PSNR) to measure reconstruction fidelity, and Structural Similarity Index Measure (SSIM) to assess perceptual quality.

\subsection{Experimental Results}
Table~\ref{tbl:main_results} summarizes reconstruction fidelity across six Sentinel-2 scenes for both Level-0 and Level-1C products.
Importantly, the compared INR baselines are evaluated using substantially larger power/energy budgets than ELMZip.
Specifically, each INR consumes approximately 10$\times$ more electrical energy than ELMZip to obtain the reported results,
reflecting the higher computational cost of iterative backpropagation-based optimization.

Despite this disadvantage in resource budget, ELMZip achieves the best reconstruction quality across all scenes,
consistently outperforming all baselines in PSNR and SSIM.
This trend is especially pronounced in structurally complex regions.
For Cairo and Seoul, which contain dense man-made structures and heterogeneous textures,
ELMZip improves PSNR by 16.5\% and 21.7\%, respectively, over the strongest INR baseline, even though the INR baseline uses 10$\times$ higher energy.
We provide detailed visualization of the experimental results in Fig.~\ref{fig:main_results}. While INR baselines trained with much higher energy budgets still exhibit over-smoothing and residual artifacts in fine structures,
ELMZip reconstructs sharper edges and preserves local contrast more faithfully, producing outputs that most closely match the ground truth.

% \section{Discussion}
% ELMZip demonstrates a practical pathway to rapid onboard compression and fast preview reconstruction under tight downlink and power constraints. By formulating the image fitting process as a convex least-squares problem and transmitting only the output parameters, the method supports timely screening of acquired data at the ground station, which is beneficial for operational workflows such as preview-based quality assessment, vegetation monitoring, and urban change detection. 
% Future work will extend the evaluation to larger-scale benchmarks (e.g., multiple Level-0 and Level-1C scenes) and explore numerical solvers to improve robustness and efficiency under constrained onboard memory and compute.

\section{Discussion}
ELMZip demonstrates a practical pathway to rapid onboard compression and fast preview reconstruction under tight downlink and power constraints. By formulating the image fitting process as a convex least-squares problem and transmitting only the output parameters, the method supports timely screening of acquired data at the ground station, which is beneficial for operational workflows such as preview-based quality assessment, vegetation monitoring, and urban change detection. In addition, its low-energy fitting aligns well with the strict power budgets of small satellite platforms.
Future work will extend the evaluation to larger-scale benchmarks and explore numerical solvers to improve robustness and efficiency under constrained onboard memory and compute.

\clearpage

\small
\bibliographystyle{IEEEtranN}
\bibliography{references}

\end{document}